\documentclass[letterpaper]{article} 
\usepackage{aaai25}  
\usepackage{times}  
\usepackage{helvet}  
\usepackage{courier}  
\usepackage[hyphens]{url}  
\usepackage{graphicx} 
\usepackage{natbib}  
\usepackage{caption} 
\usepackage{algorithm}
\usepackage{algorithmic}
\usepackage{amssymb}
\usepackage{booktabs}
\usepackage{multirow}
\usepackage{colortbl} %
\usepackage{newfloat}
\usepackage{listings}
\DeclareCaptionStyle{ruled}{labelfont=normalfont,labelsep=colon,strut=off} 
\floatstyle{ruled}
\newfloat{listing}{tb}{lst}{}
\floatname{listing}{Listing}
\usepackage{amsfonts}       
\usepackage{nicefrac}       
\usepackage{microtype}      

\title{SubjectDrive: Scaling Generative Data in Autonomous Driving via Subject Control}
\author{
    Binyuan Huang\textsuperscript{\rm 1}\thanks{Equal Contribution.}\thanks{This work was done during the internship at MEGVII.},
    Yuqing Wen\textsuperscript{\rm 2}\footnotemark[1]\footnotemark[2],
    Yucheng Zhao\textsuperscript{\rm 3}\footnotemark[1],
    Yaosi Hu\textsuperscript{\rm 4}\footnotemark[1],
    Yingfei Liu\textsuperscript{\rm 3},
    Fan Jia\textsuperscript{\rm 3},\\
    Weixin Mao\textsuperscript{\rm 3},
    Tiancai Wang\textsuperscript{\rm 3}\thanks{Corresponding author.},
    Chi Zhang\textsuperscript{\rm 5},
    Chang Wen Chen\textsuperscript{\rm 4},
    Zhenzhong Chen\textsuperscript{\rm 1},
    Xiangyu Zhang\textsuperscript{\rm 3}
}

\affiliations{
    \textsuperscript{\rm 1}Wuhan University\\
    \textsuperscript{\rm 2}University of Science and Technology of China\\
    \textsuperscript{\rm 3}MEGVII Technology\\
    \textsuperscript{\rm 4}The Hong Kong Polytechnic University\\
    \textsuperscript{\rm 5}Mach Drive\\

}

\usepackage{bibentry}

\begin{document}

\maketitle

%

\begin{abstract}
Autonomous driving progress relies on large-scale annotated datasets. In this work, we explore the potential of generative models to produce vast quantities of freely-labeled data for autonomous driving applications and present SubjectDrive, the first model proven to scale generative data production in a way that could continuously improve autonomous driving applications. We investigate the impact of scaling up the quantity of generative data on the performance of downstream perception models and find that enhancing data diversity plays a crucial role in effectively scaling generative data production. Therefore, we have developed a novel model equipped with a subject control mechanism, which allows the generative model to leverage diverse external data sources for producing varied and useful data. Extensive evaluations confirm SubjectDrive's efficacy in generating scalable autonomous driving training data, marking a significant step toward revolutionizing data production methods in this field.

\end{abstract}

\section{Introduction}
Deep generative models \cite{rombach2022high, blattmann2023align, yu2023magvit, zhang2023adding, DBLP:conf/iclr/SongME21, DBLP:conf/nips/HoJA20} have achieved remarkable progress recently, demonstrating excellence in generating high-quality and realistic visual content. Diffusion models\cite{DBLP:conf/iclr/SongME21, DBLP:conf/nips/HoJA20}, a key contributor to this advancement, are renowned for their stable and top-quality sample generation. Through the utilization of modern diffusion models, the recent breakthroughs in controllable technology \cite{zhang2023adding} now facilitate precise and flexible content customization. These developments enable the creation of synthetic samples that are almost indistinguishable from real, human-annotated data. By leveraging the generative data, methods dependent on supervised training can be substantially enhanced for tasks such as image classification\cite{imgclass}, object detection\cite{detdiffusion}, and object tracking\cite{trackdiffusion}. Motivated by these successes, there is a growing interest in applying generative models to the synthesis of natural data for more  discriminative tasks \cite{DBLP:conf/cvpr/SariyildizALK23, DBLP:journals/corr/abs-2304-08466, DBLP:conf/nips/0001VTN23}.

 \begin{figure}[!h]
\centering
\includegraphics[width=0.95\columnwidth]{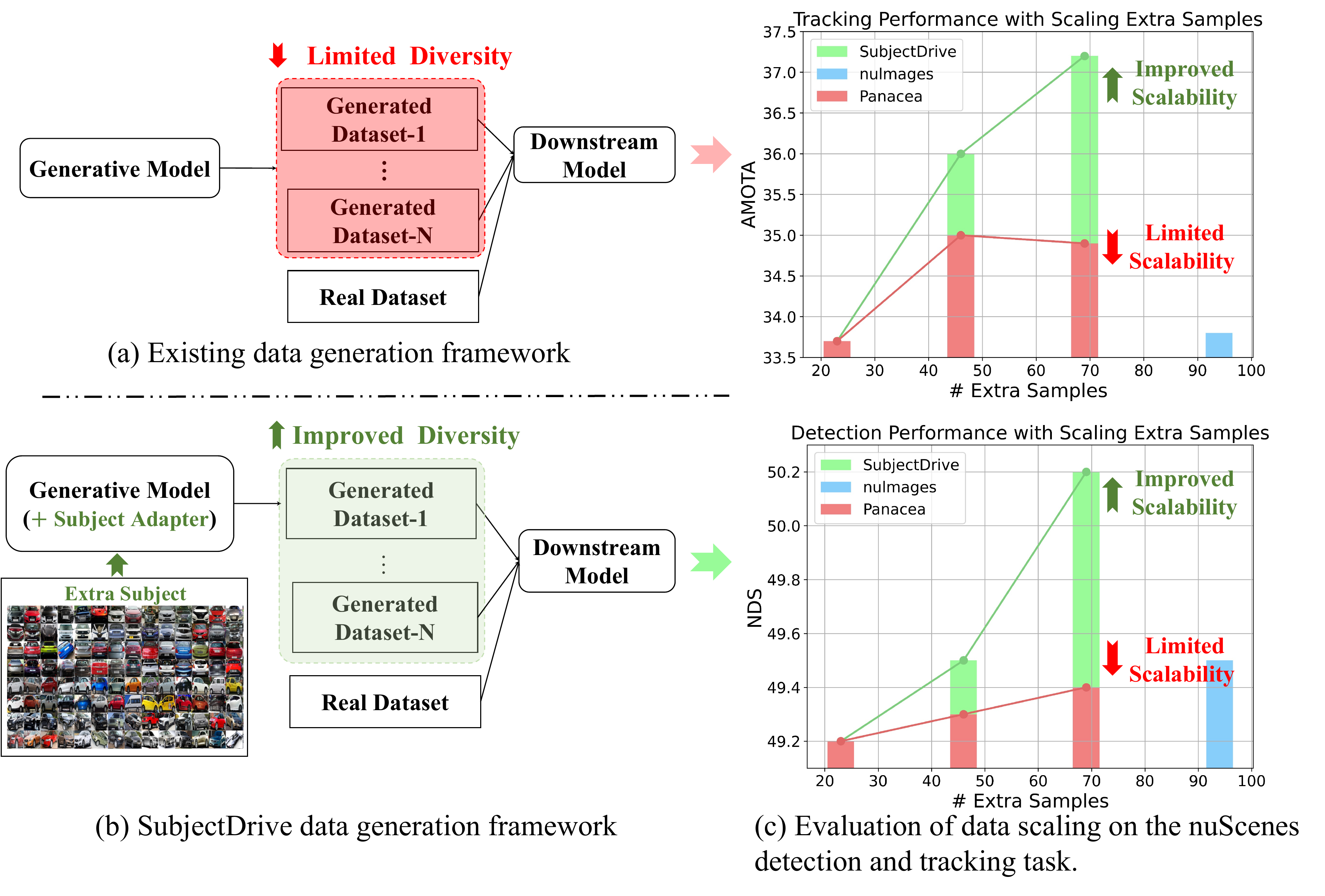}
\caption{ The comparison between  existing method and SubjectDrive framework.(a) Existing data generation framework that uses the control sequence and sampling noise to generate synthetic data with limited sample diversity and scalability. (b) Compared with the traditional framework, SubjectDrive introduces additional synthesis diversity by incorporating extra subject control to enhance the scalability of generative model.
(c) Evaluation of data scaling on the nuScenes detection and tracking task.
}
\label{fig:pipeline}
\end{figure}

Autonomous driving, as a key application domain, heavily relies on large-scale annotated data and has increasingly paid attention to generative data collection methods. One of the most challenging tasks in autonomous driving is BEV (Bird's Eye View) perception, which includes 3D object detection\cite{DBLP:journals/corr/abs-2303-11926, pang2023standing, DBLP:conf/eccv/LiWLXSLQD22, liu2022petr, liu2023petrv2}, 3D object tracking\cite{pang2023standing, fischer2022cc, marinello2022triplettrack}, and 3D lane detection. Acquiring annotated training data for these tasks is not only costly but also burdened by significant concerns regarding data privacy and usage rights, creating barriers that hinder data collection, labeling, release, and exchange processes, ultimately slowing progress in the field. To mitigate this issue, previous work has developed generative solutions such as BEVGen \cite{DBLP:journals/corr/abs-2301-04634} and BEVControl \cite{DBLP:journals/corr/abs-2308-01661}, which could produce annotated training samples by adapting an image synthesis model. More recent efforts \cite{DBLP:journals/corr/abs-2310-02601, DBLP:journals/corr/abs-2311-16813, li2023drivingdiffusion, DBLP:journals/corr/abs-2312-02934} have ventured into crafting driving scene videos, which successfully bolster more sophisticated temporal BEV perception models \cite{DBLP:journals/corr/abs-2303-11926}.

Albeit promising, most existing work focuses on exploring the potential of generative data on a small scale, which fails to fully exploit the capacity of generative models to produce an inexhaustible supply of training samples. Notably, although early methods have demonstrated superior performance when incorporating generated samples alongside real data \cite{DBLP:journals/corr/abs-2311-16813}, there is still a gap compared to using existing large-scale real image data \cite{DBLP:conf/cvpr/CaesarBLVLXKPBB20} for pre-training, as illustrated in Fig. 1(a).

To further ignite the potential of synthetic data, this paper will first investigate the scalability of generative data produced by existing methods. As shown in Fig. 1(a), we observe that when the number of synthetic samples generated by the existing method increased from 46K to 69K, the performance in 3D object tracking notably decreased. We found that this lack of scalability is intrinsically related to \textbf{the limited diversity of the generated samples}. As shown in Fig. 2, different generative samples with the same annotation exhibit remarkably similar appearances in their foreground objects, i.e., the car, despite differences in the background. We also found that when the diversity issue is addressed by the method we introduce later, the scalability of generative data can be significantly improved.

In this paper, we tackle the challenge of scaling generative data for autonomous driving. We present SubjectDrive, an advanced video generation framework designed to enhance the scalability of generative models. Our initial findings reveal that conventional video generation pipelines struggle to scale effectively with increased data volumes due to its limited diversity. To overcome this limitation, we propose a novel generation framework centered on augmenting sampling diversity. Specifically, we integrate a feature termed subject control into existing generation pipelines. This feature empowers generative models to manipulate the diversity of the synthesis process by providing a mechanism to dictate the visual appearance of foreground elements in generated samples. This innovative feature enables the blending of the inherent stochastic nature of the generative process with the diversity drawn from external data sources, thereby crafting a more powerful model capable of producing scalable and diverse samples.

\begin{figure}[!t]
\centering
\includegraphics[width= 0.85\columnwidth]{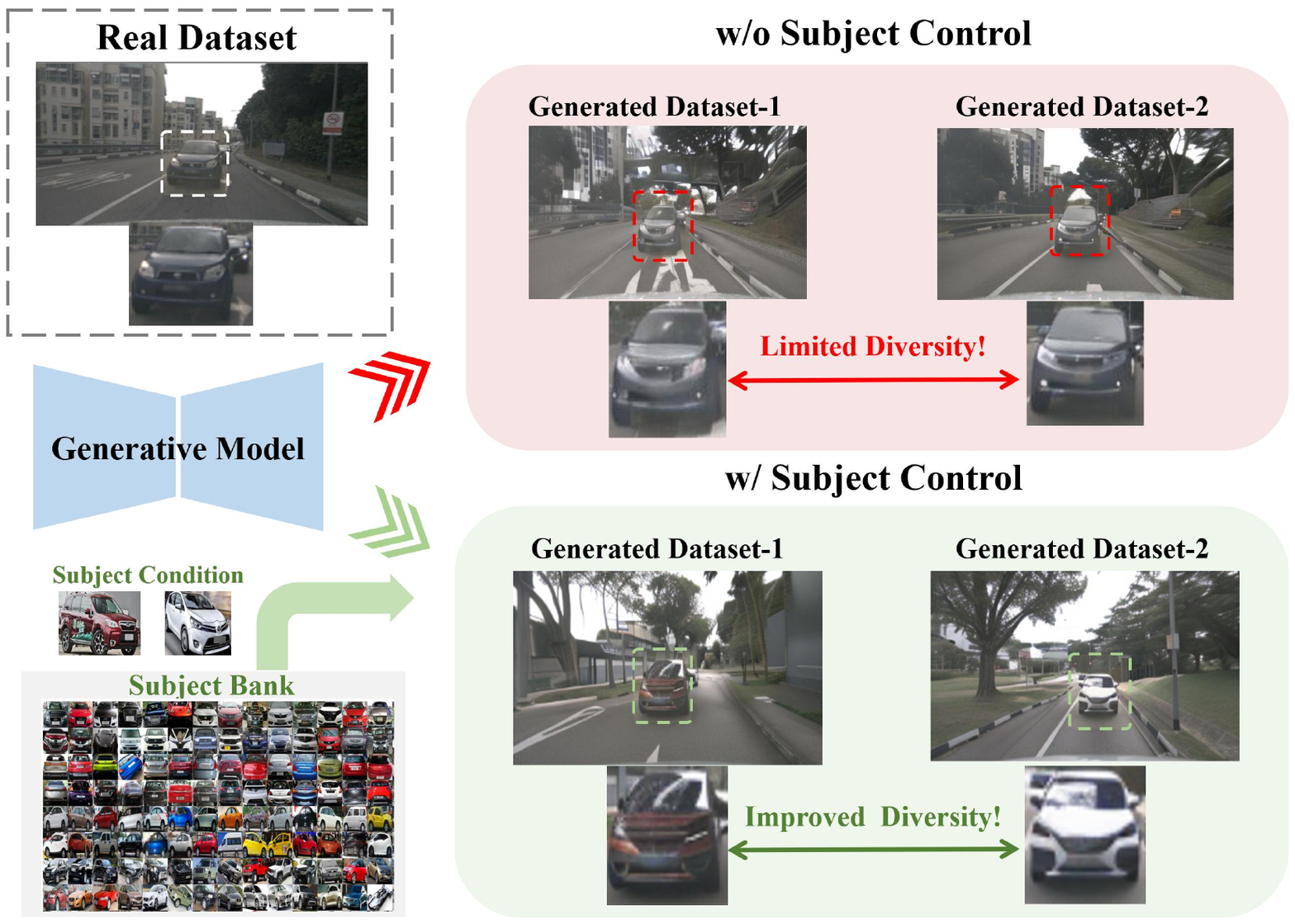}
\caption{
Above: Existing autonomous driving generative models struggle to produce diverse foreground samples.
Below: By enhancing the sampling diversity capabilities with our subject control methods, the diversity of the generated foreground sample has significantly improved.
}
\label{fig:motivation}
\end{figure}

More concretely, SubjectDrive is designed around a trio of innovative modules that collectively enable robust subject control capabilities. Initially, the model leverages a \textbf{Subject Prompt Adapter}, integrating subject control with the existing text-conditioned branch. Subsequently, to enhance the model's ability to inject spatial information, we introduce a \textbf{Subject Visual Adapter} that directly utilizes visual features, incorporating them into the existing diffusion U-Net architecture. Lastly, to ensure consistent injection of these features over time, we deploy \textbf{Augmented Temporal Attention} that expands the model's temporal-spatial context.  Together, these modules empower SubjectDrive to perform subject-conditioned video generation with compelling efficacy. Overall, our research includes three main contributions:

\begin{itemize}
\item Unlike existing synthetic methods for autonomous driving, which are limited to small-scale validations, we propose a novel and critical research direction: investigating the scalability of generative data in autonomous driving. Our exploration provides a deeper understanding of the challenges and solutions associated with leveraging large-scale synthetic data in this domain.

\item  To address the challenges faced by existing methods during data scaling, particularly the issue of limited diversity, we introduce a novel generative framework called \textit{SubjectDrive}. This framework enhances sampling diversity and improves the scalability of generative data through an innovative mechanism known as subject control.

\item Extensive experiments on the nuScenes dataset  \cite{DBLP:conf/cvpr/CaesarBLVLXKPBB20} validate the effectiveness of our proposed method. Remarkably, our approach is the first generative technique to surpass the performance of perception models pre-trained on a large-scale real dataset, specifically nuImages\cite{DBLP:conf/cvpr/CaesarBLVLXKPBB20}. These outstanding results underscore the transformative potential of generative data in advancing autonomous driving technologies, offering a promising direction for future research in this field.
\end{itemize}

\section{Related Work}

\subsection{Scalable Data Synthesis for Autonomous Driving}

The use of synthetic data has been widely explored for visual perception tasks that require great of efforts in labeled data collection. Examples include applications in image classification \cite{DBLP:journals/corr/abs-2304-08466, DBLP:conf/cvpr/SariyildizALK23}, semantic segmentation \cite{DBLP:conf/nips/0001VTN23}, and object tracking \cite{DBLP:journals/corr/abs-2312-00651}. Recently, there has been a burgeoning interest in employing generative data for the challenging BEV perception tasks \cite{DBLP:journals/corr/abs-2303-11926, DBLP:conf/eccv/LiWLXSLQD22}, which demand precise geometric and appearance alignment from generative models. Initial works aim at synthesizing street-view images \cite{DBLP:journals/corr/abs-2301-04634, DBLP:journals/corr/abs-2308-01661}, with subsequent studies extending into the generation of driving scene videos \cite{DBLP:journals/corr/abs-2311-16813, DBLP:journals/corr/abs-2312-02934, DBLP:journals/corr/abs-2309-09777}. Despite these successful efforts, the majority of existing research has been limited to producing a small quantity of training samples, not fully exploiting the generative models' capacity to offer an unlimited reservoir of samples. Data scaling efforts \cite{DBLP:journals/corr/abs-2304-08466} are sparse and, thus far, not particularly successful. It's imperative to recognize that recent breakthroughs highlight the paramount importance of data scaling in training sophisticated deep learning models, such as GPT \cite{DBLP:journals/corr/abs-2303-08774}, SVD \cite{DBLP:journals/corr/abs-2311-15127}, and Sora \cite{sora}. Therefore, addressing the challenge of amplifying generative data volumes is of essence. This work delves into scalable data synthesis for autonomous driving, providing analysis results and proposing an enhanced framework.

\subsection{Diffusion-based Generative Models with Subject Controls}

Diffusion models with subject control are designed to embed target subjects into generated visual content, guided by reference images \cite{DBLP:journals/corr/abs-2305-15779}. These approaches \cite{DBLP:conf/iccv/HanLZMMY23, DBLP:conf/cvpr/KumariZ0SZ23, DBLP:journals/corr/abs-2306-17154, DBLP:journals/corr/abs-2307-09481, DBLP:journals/corr/abs-2307-11410,fastcomposer} initially emerged in image generation research to facilitate identity-preserving applications. Early methods \cite{DBLP:conf/iccv/HanLZMMY23, DBLP:conf/cvpr/KumariZ0SZ23, DBLP:journals/corr/abs-2306-17154,DBLP:conf/iclr/GalAAPBCC23, DBLP:conf/cvpr/RuizLJPRA23} adopted a fine-tuning framework to adjust a specific model capable of generating images of the desired subject. A notable example is DreamBooth \cite{DBLP:conf/cvpr/RuizLJPRA23}, which fine-tunes a diffusion model using a small set of subject images. While these fine-tuning methods achieved high-quality results, they were limited by the significant resources required for model tuning, making them impractical for scenarios with numerous target entities. To address these limitations, tuning-free methods \cite{DBLP:journals/corr/abs-2307-11410, DBLP:journals/corr/abs-2308-06721} were developed. Subject Diffusion \cite{DBLP:journals/corr/abs-2307-11410}, for instance, achieved impressive customization by introducing a subject conditioning module and training on a bespoke dataset of subject pairs. Subsequent works, such as Cones2 \cite{DBLP:journals/corr/abs-2305-19327} and others \cite{DBLP:journals/corr/abs-2307-09481}, have expanded tuning-free subject control to scenarios involving multiple subjects. 

In the video domain, there is also growing interest in subject-controllable generation \cite{DBLP:journals/corr/abs-2312-00777, DBLP:journals/corr/abs-2311-00990, DBLP:journals/corr/abs-2401-09962}. For instance, VideoBooth \cite{DBLP:journals/corr/abs-2312-00777} proposed a framework capable of generating consistent videos containing the subjects specified in image prompts. 
Additionally, CustomVideo \cite{DBLP:journals/corr/abs-2401-09962} introduced a multi-subject-driven text-to-video model powered by a simple yet effective co-occurrence and attention control mechanism.
Our work falls into the category of subject-driven video generation. However, unlike previous efforts, we explore the significance of such subject-controlled generation for data scaling, especially in the context of autonomous driving.

\section{Method}

\begin{figure*}[!t]
  \centering
   \includegraphics[width=0.85\linewidth]{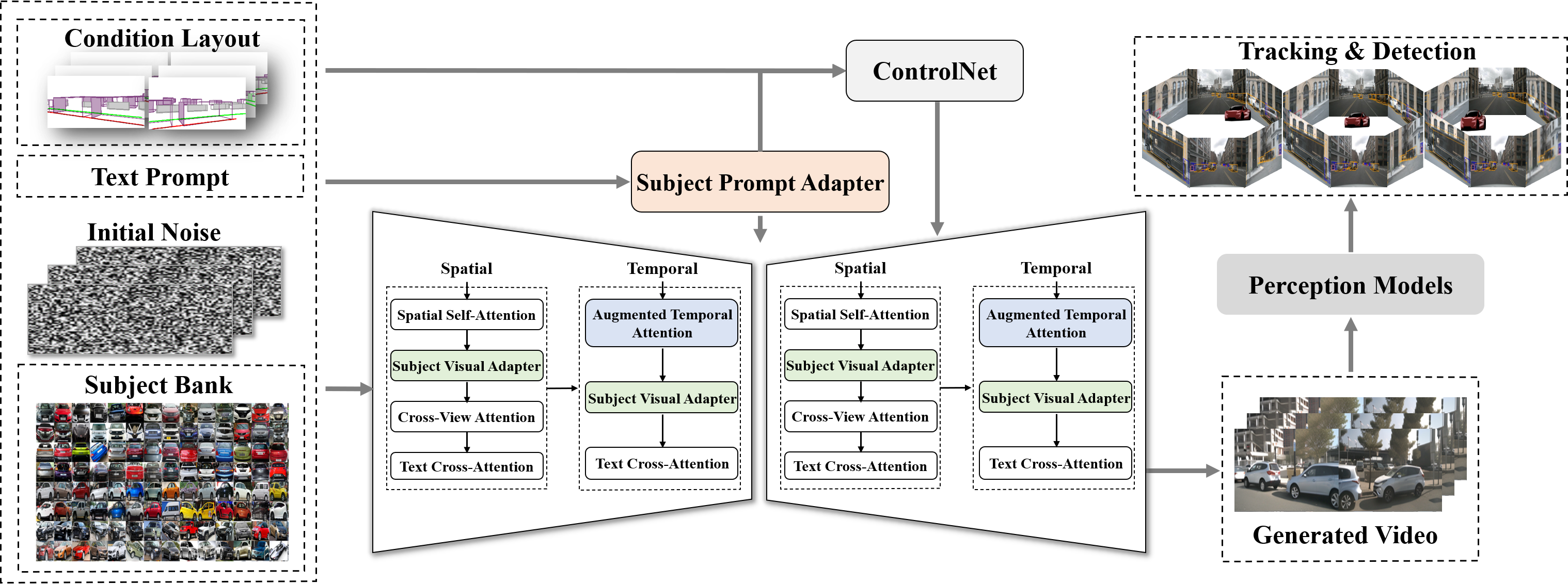}
    \caption{
    Overview of SubjectDrive.
    The pipeline of SubjectDrive involves a frozen auto-encoder and a trainable UNet-based diffusion model. Different control signal sources including extended text prompt, condition layout, and subject bank.
   }
   \label{fig:train}
\end{figure*}

\subsection{Preliminary: Latent Diffusion Models}
Diffusion Models (DMs) \cite{DBLP:conf/iclr/SongME21, DBLP:conf/nips/HoJA20} iteratively denoise a random noise with Gaussion distribution  $\mathbf{\epsilon} \sim \mathcal{N}(\mathbf{0}, \mathbf{I})$ over T steps to generate target data according to Eq. \ref{eq:noise_reduction}, where the functions $\mu_{\theta}$ and $\Sigma_{\theta}$ are derived from the denoising model $\epsilon_{\theta}$. 
The training of DMs includes both a diffusion process and a denoising process. The diffusion process begins by adding Gaussion noise to the original data sample $x_0$ over $t$ steps, controlled by the scheduled noise strength $\beta_t$, which can be simplified as Eq. \ref{eq:forward_noise_mix} where $\bar{\alpha}_t=\prod_{i=1}^t\left(1-\beta_t\right)$. To learn the denoising process, the denoising model $\mathbf{\epsilon}_{\theta}$, which aims to estimate the original noise $\mathbf{\epsilon}$ from the noisy data $x_t$, is optimized by minimizing the loss function as shown in Eq.~\ref{eq:loss_function}. 
\begin{equation}
\label{eq:noise_reduction}
p_{\theta}(x_{t-1} | x_{t}) = \mathcal{N}(\mathbf{x}_{t-1} ; \mathbf{\mu}_{\theta}(\mathbf{x}_{t}, t), \mathbf{\Sigma}_{\theta}(\mathbf{x}_{t}, t))
\end{equation}
\begin{equation}
\label{eq:forward_noise_mix}
\mathbf{x}_{t} = \sqrt{\bar{\alpha}_t} \mathbf{x}_0 + \sqrt{1-\bar{\alpha}_t} \mathbf{\epsilon}, \mathbf{\epsilon} \sim \mathcal{N}(\mathbf{0}, \mathbf{I}), \mathbf{x}_0 \sim p(x)
\end{equation}
\begin{equation}
\label{eq:loss_function}
\min_{\theta}\mathbb{E}_{t, x, \epsilon}\|\mathbf{\epsilon} - \mathbf{\epsilon}_{\theta}(x_{t}, t)\|^2
\end{equation}

Given the significant computational burden of diffusion models in generating high-resolution images or videos, Latent Diffusion Models (LDMs) \cite{rombach2022high} have become popular in recent studies. With a pre-trained auto-encoder \cite{kingma2013auto} to handle the compression and reconstruction between high-dimensional visual data and low-redundancy latent representations, the diffusion model can concentrate exclusively on generation in latent space, significantly reducing computational costs. Consequently, our method employs LDMs for video generation.

\subsection{SubjectDrive}
SubjectDrive is designed to enhance the scalability of generative data, thereby promoting perception models for autonomous driving applications.
Despite the ability of advanced generative methods \cite{DBLP:journals/corr/abs-2311-16813} to produce high-quality driving-scene videos, they narrowly uplifts the performance on downstream perception tasks as illustrated in Fig. \ref{fig:pipeline}(c). We believe this is primarily due to the limited diversity of generated foreground elements, which are crucial for autonomous driving. Thus, different from tradition generation pipeline using control sequence to guide global scene generation, we innovatively integrate subject control mechanism into generation process, allowing for the injection of external subjects from extensive open-source data. The integration of controlled subjects not only boosts controllability but also effectively enhances the diversity of generated foreground elements, which shows strong augmentation for autonomous driving applications.

To inject subjects into generation process, we propose the novel Subject Prompt Adapter (SPA) and Subject Visual Adapter (SVA) to augment expressity of text embedding with regard to subjects and integrate subjects' spatial information into frames, respectively. To further improve the appearance consistency of injected subjects across frames, the Augmented Temporal Attention (ATA) is introduced to effectively capture long-range movements in driving videos.

\subsubsection{Overview}
Our framework is built on a text-to-video diffusion model, i.e. Panacea \cite{DBLP:journals/corr/abs-2311-16813}, which is a strong baseline for multi-view video generation. 
The overall training framework of SubjectDrive is illustrated in Fig.~\ref{fig:train}. The diffused noisy input is fed into the trainable diffusion model to generate latent video under the guidance of text, condition layout, and injected subjects. We inherit the guidance of condition layout with ControlNet \cite{zhang2023adding}  from Panacea. For subjects guidance, we first extend text prompt and augment the text embedding of subject part with Subject Prompt Adapter. On the other hand, we insert the gated self-attention layer into diffusion model to enhance the location guidance of subjects captured by Subject Visual Adapter. Furthermore, we replace the conventional temporal 1D attention with proposed Augmented Temporal Attention to improve the temporal consistency of injected subjects.

\subsubsection{Subject Prompt Adapter}\label{sec:prompt_adpter}
Subject Prompt Adapter introduces an enhancement to textual cues through the integration of subject attributes including the category semantic, ID identifier, and visual semantic information. 

\begin{figure}[!h]
  \centering
\includegraphics[width=0.725\linewidth]{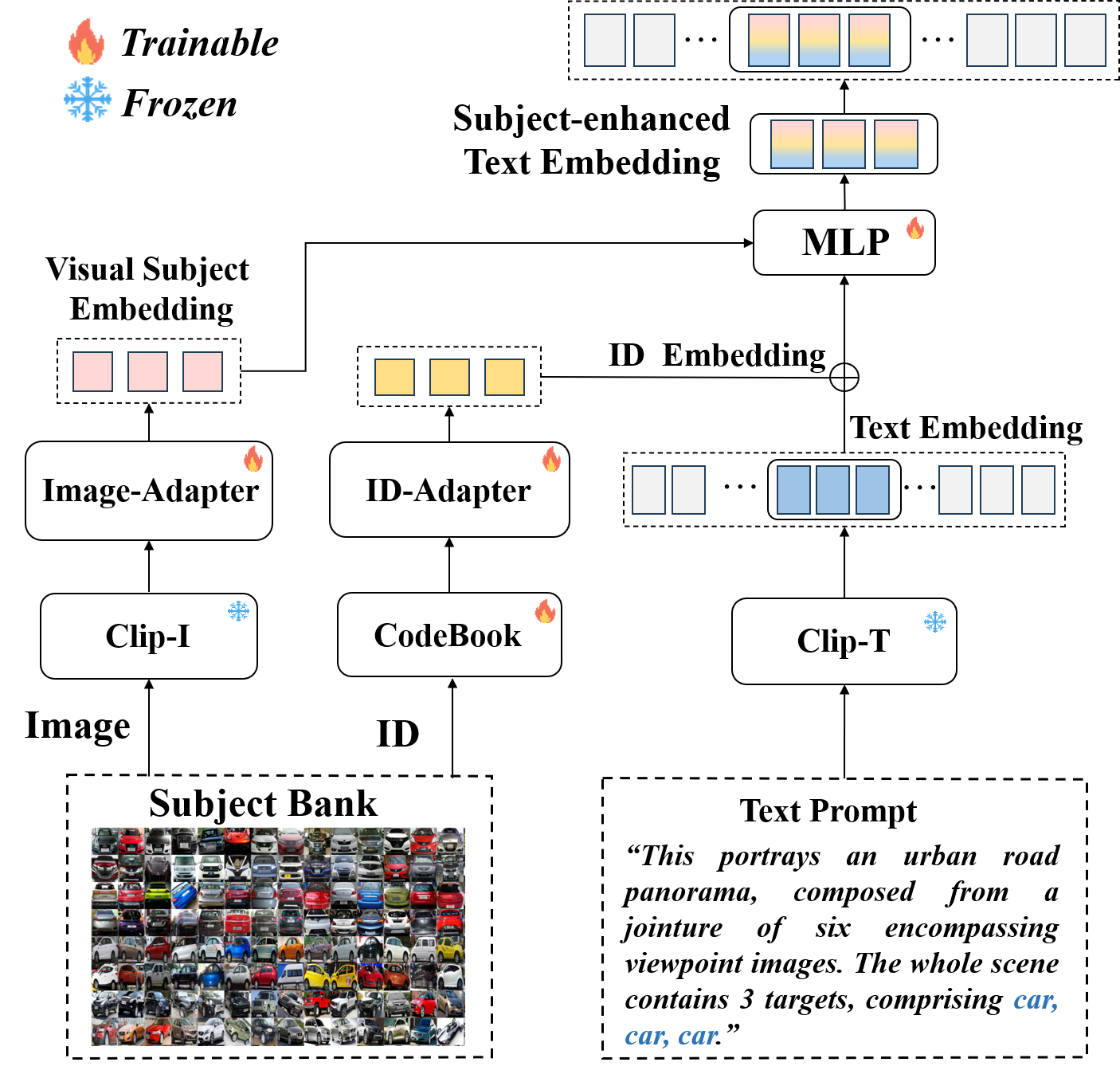}
   \caption{The Subject Prompt Adaptor which augments the original text embedding of extended prompt with corresponding ID identifier and visual semantic information to enhance the expressity of subjects. }
   \label{fig:subject}
\end{figure}

To achieve this, we first extend the scene prompt with injected subject category description.
For example, the prompt for the $N_{th}$ frame that contains $M$ subjects is organised as
"\textit{[Scene Prompt], including [Subject $X_1$], [Subject $X_{2}$].... [Subject $X_{M}$].}"
\textit{[Subject $X_i$]} here refers to the category word of subject $X_i$, e.g., car, bus.
The extended prompt is input into the pre-trained CLIP \cite{radford2021learning} text encoder to obtain the original text embedding $z^{t} \in {\mathbb{R} ^{L \times d}}$, where $L$ represents the length of the text embedding and $d$ is the dimension of the text embedding.
To further integrate subject attributes for better expressity, we extract the ID identifier of each subject from condition layout to ensure coherent trajectories across frames, and encode it with a learnable codebook and an ID-adapter comprised of MLP to get the ID embedding  $z_{i}^{id} \in {\mathbb{R} ^{d}}$ for subject $X_i$. Similarly, we encode the image of $X_i$ by the CLIP image encoder and an image-adapter to obtain the corresponding visual semantic embeddings $z_{i}^{v} \in {\mathbb{R} ^{d}}$.

For the original text embedding $z_{i}^{t}$ at the index of subject $X_i$ in $z^{t}$, we successively enhance it with the corresponding ID embeddings  $z_{i}^{id}$ and visual semantic embeddings $z_{i}^{v}$ by
\begin{equation}
   \hat{z}_{i}^{t} = \mathrm{MLP} \left( [z_{i}^{t} + z_{i}^{id}, z_{i}^{v}] \right), \ \ i \in\left\{\mathrm { Index }_{1,2, \cdots, M}\right\}
\end{equation}
where $[,]$ represents the concatenation operation. The resulted subject-enhanced text embedding $\hat{z}^{t}$ contains not only the semantic of whole scene, but also specific identifying information for each subject. The subject-enhanced text embedding is further injected into the UNet through the cross-attention layer to guide the generation of frames.

\subsubsection{Subject Visual Adapter}\label{sec:visual_adapter}
The Subject Visual Adapter is proposed to inject subject spatial information into video feature to further enhance content alignment of generated video with provided visual clue and location of subjects.

\begin{figure}[!h]
  \centering
\includegraphics[width=0.8\linewidth]{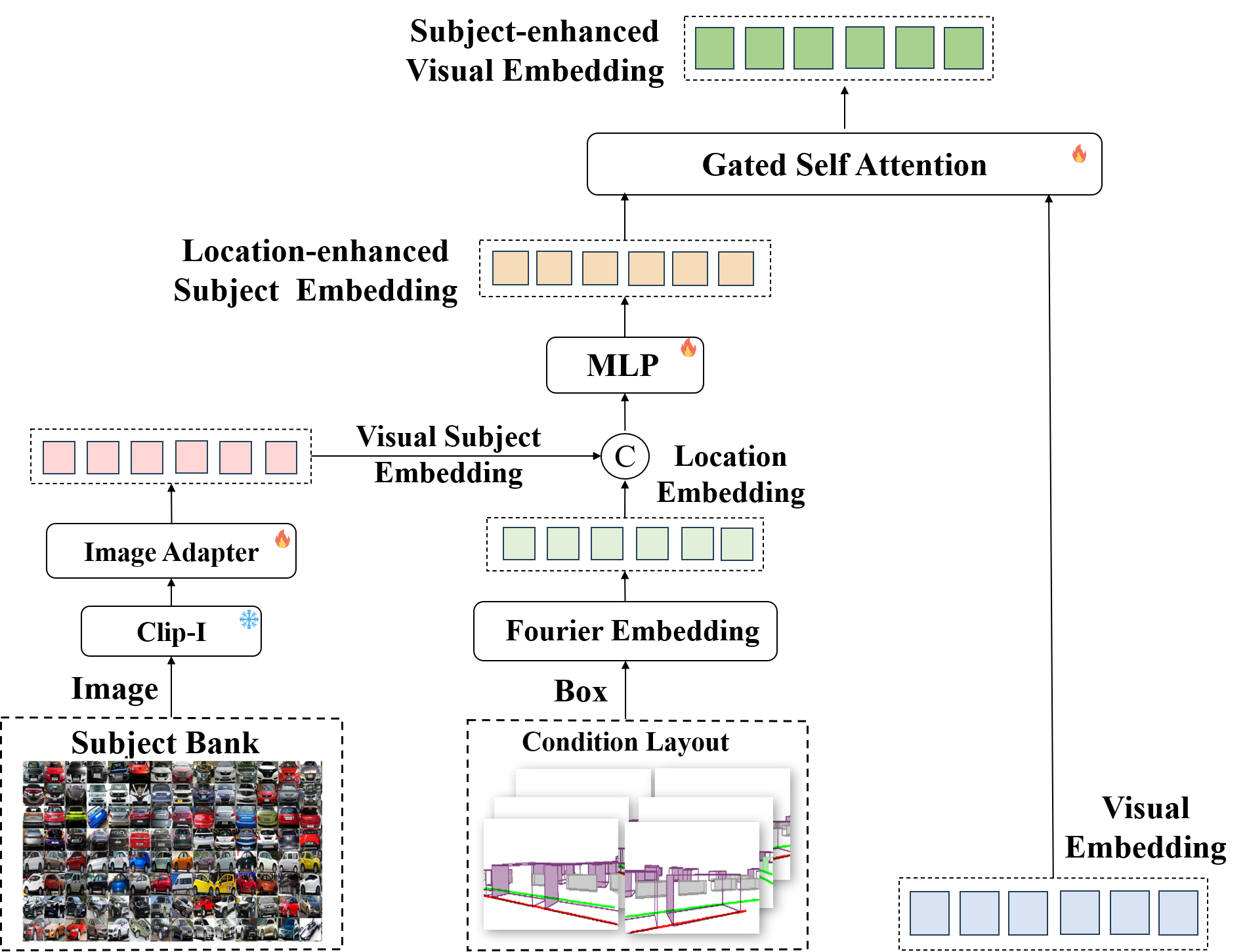}
   \hfill
   \caption{The Subject Visual Adapter which injects location-enhanced subject information into visual features cooperated with gated self-attention.}

   \label{fig:subject}
   
\end{figure}
Inspired by \cite{DBLP:conf/cvpr/LiLWMYGLL23}, SVA first combines subject images and corresponding locations to form location-enhanced subject embeddings, and then injects them into frames with a control signal to guide the generation of subjects at specified location.
To obtain the location-enhanced subject embedding $f^{vl}$, 
the location of each subject, represented by its coordinates  $l=[x_{min}, y_{min}, x_{max}, y_{max}]$ in the current frame, is encoded with fourier embedding, further integrated into visual embedding $f^{v}$ of corresponding subject as
\begin{equation}
f^{vl} = \mathrm{MLP}[f^{v}, \mathrm{Fourier}(l)]).
\end{equation}

To inject the location-enhanced subject embedding into generated frame, we employ the gated self-attention layer \cite{DBLP:conf/cvpr/LiLWMYGLL23} that locates between each paired self-attention layer and cross-attention layer in the UNet. In the gated self-attention layer, the location-enhanced subject embedding $f^{vl}$ interacts with frame visual tokens $z$ by an attention operation to capture the dependency between subject embedding and visual tokens, followed by a token selection operation $TS$ to preserve only visual tokens. To adaptively adjust the guidance scale of location information over frames, a gating factor is learned which is operated as 
\begin{equation}
z = z +  \tanh(\gamma) \cdot \mathrm{TS}(\mathrm{SelfAttn}([z, f^{vl}])),
\end{equation}
where $\gamma$ represents the gating factor, a learnable parameter initialized to 0 for stable training.

\subsubsection{Augmented Temporal Attention}\label{sec:motion_module}
The Augmented Temporal Attention is designed to  capture long-range movement of subjects with feasible computation cost.
\begin{figure}[!h]
  \centering
\includegraphics[width=0.9\linewidth]{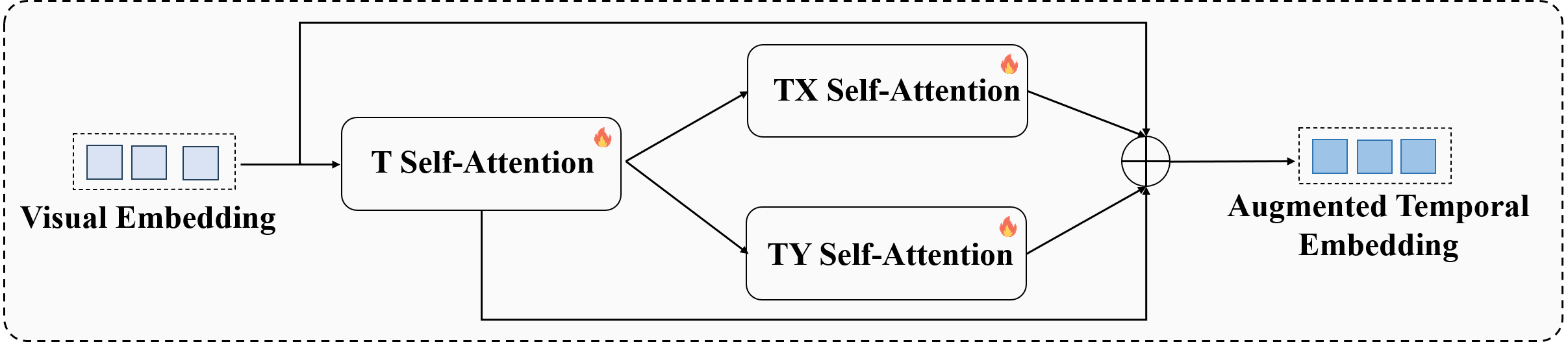}
   \hfill
   \caption{The
Augmented Temporal Attention integrates conventional temporal 1D attention with
decomposed attention on temporal-horizontal (TX) plane and temporal-vertical (TY)
plane to effectively capture long-range movements of subjects.}
   \label{fig:subject}
\end{figure}
The conventional temporal attention layer utilized in video diffusion models solely incorporates self-attention operation within the temporal dimension. However, due to the substantial movements typically involved with subjects in autonomous driving videos, it becomes challenging for the temporal-dimension attention to effectively capture the long-range dependency of inter-frame subjects. Therefore, we propose the Augmented Temporal Attention which incorporates interaction within temporal-horizontal (TX) plane and temporal-vertical (TY) plane to improve subject consistency across sequences.

Specifically, given input video feature $z^{i} \in R^{T\times H\times W \times C}$ where $T$, $H$, $W$, and $C$ denote the video length, spatial height, width, and number of channels, respectively, it is processed by 1D attention along temporal dimension (T) to obtain $z^{T}$.  
$z^{T}$ is reshaped to respective $z^{TX} \in R^{H\times \left( T\times W\right) \times C}$ and $z^{TY} \in R^{W\times \left( T\times H\right) \times C}$ to aggregate information from two decomposed planes by the parallel TX self-attention and TY self-attention.
In this way, the fused video feature not only integrates small-range variation from temporal 1D attention, but also captures large-range movements by decomposed TX and TY attention, which can effectively enhance the temporal coherence of the generated video. The overall operation of this layer can be written as

\begin{equation}
z^{o} = \mathrm{SelfAttn}(z^{TX}) + \mathrm{SelfAttn}(z^{TY}) +  z^{T} + z^{i}.
\end{equation}

\section{Experiment}

\subsection{Evaluation Datasets and Metrics}

\noindent\textbf{Datasets.}
We employ the nuScenes dataset to train SubjectDrive and utilize it to evaluate the visual fidelity and controllability of the generated data. The nuScenes dataset comprises 1,000 scenes, each lasting 20 seconds, with annotations provided at a frequency of 2Hz and featuring a comprehensive 360° camera field of view. It includes 1.4 million 3D bounding boxes, spanning 10 categories: car, truck, bus, trailer, construction vehicle, pedestrian, motorcycle, bicycle, barrier, and traffic cone.

\noindent\textbf{Evaluation Metrics.} Following Panacea, we evaluate two key objectives using perceptual models. First, we assess the generative model's controllability by measuring how well the generated data aligns with BEV annotations, using a pre-trained StreamPETR model to compare performance against real data. Second, we evaluate how the generated data improves perceptual model performance on detection and tracking tasks. Detection metrics include nuScenes Detection Score (NDS), mean Average Precision (mAP), mean Average Orientation Error (mAOE), and mean Average Velocity Error (mAVE). Tracking metrics include Average Multi-Object Tracking Accuracy (AMOTA), Precision (AMOTP), Recall, and Accuracy (MOTA). Visual fidelity is validated with Fréchet Inception Distance (FID)\cite{DBLP:conf/nips/HeuselRUNH17} and Fréchet Video Distance (FVD)\cite{DBLP:journals/corr/abs-1812-01717}.

\subsection{Implementation Details}

SubjectDrive adopts a two-stage video generation approach: image generation in the first stage (optimized for 56k steps) and video generation in the second (84k steps). Experiments are conducted on 8A100 GPUs using the DDIM sampler with 25 steps to produce $256\times512$ resolution video clips spanning 8 frames. The evaluation uses StreamPETR with a ResNet50 backbone~\cite{he2016deep}, trained at $256\times512$ resolution.
To construct the subject bank, we draw from two distinct sources, categorizing them into internal and external subject banks. The internal subject bank is curated by collecting subjects from the training set of the nuScenes dataset. The external subject bank is established by integrating external vehicle datasets from the open-source CompCars \cite{yang2015large} dataset. 
The internal subject bank is used during the training phase, while the external subject bank is mixed with the internal one during the sampling phase to promote the generation of diverse data.

\subsection{Main Results}

\textbf{Analysis of Synthetic Data Scale-up.} We investigate the impact of scaling up the quantity of generative data on the performance of downstream tasks and present the results in Table \ref{tab:scale}. Two interesting findings emerge. First, \textbf{increasing the quantity of synthetic data has a positive influence on the performance of the perception model.} Although using a small amount of synthetic data is initially less effective than using real nuImages data, increasing the data volume can boost the models to achieve superior performance in both NDS and AMOTA. This demonstrates that scaling up generative data is essential to unlock the potential of generative data production. Second, \textbf{the integration of external subjects can substantially improve scaling performance.} For instance, while tripling the generative data from Panacea yields a mere 0.2 improvement in NDS, incorporating SubjectDrive at the same data scale elevates the NDS by 1.0. Similar trends are also observed in AMOTA. These results demonstrate that utilizing external subjects is an effective way to enhance the generative model’s scaling capability.

\begin{table}[!h]
  \centering
  \small
  \renewcommand{\arraystretch}{1.0} 
  \setlength{\tabcolsep}{1.0pt}
\begin{tabular}{llllll}
\toprule
\multirow{2}{*}{\#Extras} & \multirow{2}{*}{\hspace{1em}Source} & \multicolumn{2}{l}{\hspace{2.5em}Tracking} & \multicolumn{2}{l}{\hspace{2em}Detection} \\ \cline{3-6} 
  &            &  AMOTA $\uparrow$           & AMOTP $\downarrow$    &  NDS $\uparrow$     & MAP $\uparrow$      \\ \hline
0                                   & nuScenes                     & 30.1             & 1.379     & 46.9              & 34.5      \\ \hline
93K                              & nuImages                   & 33.8 (+3.7\%)     & 1.324     & 49.5 (+2.6\%)      & 37.8      \\ \hline
23K                               & \multirow{3}{*}{Panacea}                       & 33.7 (+3.6\%)     & 1.353     & 49.2 (+2.3\%)      & 37.1      \\
46K                               &                 & 35.0 (+4.9\%)     & 1.346     & 49.3 (+2.4\%)      & 37.3      \\
69K                               &                & 34.9 (+4.8\%)     & 1.348     & 49.4 (+2.5\%)      & 37.1      \\ \hline
23K                               & \multirow{3}{*}{Ours}              & 33.7 (+3.6\%)     & 1.353     & 49.2 (+2.3\%)      & 37.1      \\ 
46K                               &               & 36.0 (+5.9\%)     & 1.322     & 49.5 (+2.6\%)      & 37.7      \\ 
69K                               &                & \textbf{37.2(+7.1\%)}    &\textbf{ 1.317}      & \textbf{50.2(+3.3\%)}      &\textbf{38.1}        \\ \bottomrule
\end{tabular}
\caption{Evaluation of data scaling on detection and tracking tasks. Extras denote extra samples beyond nuScenes dataset.}
\label{tab:scale}
\end{table}

\noindent\textbf{Analysis of Performance in 3D Object Detection Task.} Next, we present a quantitative analysis that compares SubjectDrive with other data generation methods in the 3D detection task. We utilize SubjectDrive to generate novel synthetic data as additional training source for the StreamPETR model, and then evaluate its performance on the real nuScenes validation set. The results are showcased in Table \ref{tab:detection}. When trained solely on the synthetic data, SubjectDrive manages to outperform Panacea by 5.5 mAP and 5.0 NDS. Specifically, SubjectDrive achieves an NDS of 41.1, reaching 88\% of the performance compared to that trained with real nuScenes data. When combining synthetic data with real data, SubjectDrive surpasses all other methods by a significant margin.

\begin{table}[h]
    \centering
    \small
\renewcommand{\arraystretch}{1.0} 
\setlength{\tabcolsep}{1.25pt}
\begin{tabular}{llccc}
\toprule
DataType & Method & MAP$\uparrow$ & NDS$\uparrow$ & MAVE$\downarrow$ \\ \hline
Real & Real Only & 34.5 & 46.9 & 29.1  \\ \hline                   
\multirow{2}{*}{Gen} & Panacea
 & 22.5 & 36.1 & 46.9 \\
& Ours & \textbf{28.0 (+5.5\%)} & \textbf{41.1 (+5.0\%)}  & \textbf{37.0}\\ \hline
\multirow{4}{*}{Real+Gen} 
& DriveDreamer & 35.8 & 39.5 & - \\ 
& WoVoGen$*$  & 36.2 & 18.1 &123.4 \\ 
& MagicDrive  & 35.4 & 39.8  & -\\ 
& Panacea  & 37.1 & 49.2 &27.3 \\ 
& Ours & \textbf{38.1} & \textbf{50.2} &\textbf{26.4}  \\
\bottomrule
\end{tabular}
\caption{Comparison on the 3D object detection task with other generation methods. $*$ indicates the evaluation of WoVoGen is only on the vehicle classes of cars, trucks, and buses.}
\label{tab:detection}
\end{table}

\noindent\textbf{Analysis of Performance in 3D Object Tracking Task.} In addition to the 3D object detection task, the 3D object tracking task is an important and more challenging fundamental task in autonomous driving. As shown in Table \ref{tab:tracking}, using solely our synthetic data can produce a model with a 23.5 MOTA, attaining 86\% of the performance compared to that trained with real nuScenes data. When combining synthetic data with real data, SubjectDrive achieves a 37.2 AMOTA, which is 3.5 points higher than the model trained with Panacea. It is worth noting that the performance gain in tracking is notably higher than that in detection. This is because our subject control mechanism not only improves data diversity but also has a beneficial side effect of enhancing temporal coherency by ensuring all generated objects across frames align with the given reference subjects.

\begin{table}[!h]
\centering
\small
\footnotesize
\renewcommand{\arraystretch}{1.02} 
\setlength{\tabcolsep}{1mm}
\begin{tabular}{lcccc}
\toprule
DataType & Method & AMOTA $\uparrow$   & AMOTP $\downarrow$ & MOTA $\uparrow$ \\ \hline
Real & Real Only & 30.1 & 1.379 & 27.1  \\ \hline
Gen & Ours & 23.4 & 1.544  & 23.5\\ \hline
\multirow{2}{*}{Real+Gen} 
& Panacea & 33.7  & 1.353  & 30.6\\
& Ours & \textbf{37.2 (+3.5\%)}   & \textbf{1.317}  & \textbf{33.3 (+2.7\%)}\\ \hline
\end{tabular}
\caption{Comparison on the 3D object tracking task.}
\label{tab:tracking}
\end{table}

\noindent\textbf{Comparison of Generation Quality.} In order to validate the visual quality of our generated samples, we compared them with state-of-the-art methods for driving scene generation. We generated the validation set of nuScenes without applying any pre-processing or post-processing to the selected samples. As shown in Table \ref{tab:fvd}, our method achieves the best performance, with an FVD of 124 and an FID of 15.98, compared to both video-level generation methods—WoVoGen, Panacea, and DriveDreamer—and image-level generation methods, BEVGen , BEVControl and MagicDrive.

\begin{table}[h]
    \centering
    \renewcommand{\arraystretch}{1.0} 
    \small
        \centering
        \begin{tabular}{lcccc}
            \toprule
            Method     &Multi-View &Multi-Frame  & FVD$\downarrow$ & FID$\downarrow$ \\
            \hline
            BEVGen \  &$\checkmark$   &  & - & 25.54\\
            BEVControl &$\checkmark$   &  & - & 24.85 \\
            MagicDrive  &$\checkmark$   &  & - & 16.20 \\
           DriveDreamer    & &$\checkmark$  & 353 & 26.8\\
           WoVoGen    &$\checkmark$  & $\checkmark$ & 418  & 27.6  \\
             Panacea &$\checkmark$  &$\checkmark$  & 139  & 16.96  \\
               \hline
              
            Ours   &$\checkmark$  &$\checkmark$  & \textbf{124}  & \textbf{15.98}  \\
            \bottomrule
        \end{tabular}
         \caption{ Comparison of FID and FVD metrics with state-of-the-art methods on the validation set of the nuScenes dataset.}
        \label{tab:fvd}
\end{table}

\subsection{Ablation Studies}
The alignment between generated samples and the provided BEV conditional labels is essential for evaluating generative models' controllability. It also acts as a key indicator of the synthetic data's applicability. In this section, we conduct ablation studies to justify the design choices in SubjectDrive using this metric.
Table \ref{tab:ablation} presents our evaluation results for the Subject Visual Adapter (SVA), Subject Prompt Adapter (SPA), and Augmented Temporal Attention (ATA). First, compared to the baseline, introducing the SVA resulted in a significant improvement of 2.2 in NDS and 4.2 in AMOTA, demonstrating its effectiveness in enhancing label alignment. This improvement is achieved by injecting extra subject control through visual features and their corresponding positional embeddings. Second, by integrating the SPA with the SVA, we observed an additional increase of 1.8 in NDS and AMOTA. These results indicate that these two subject control modules are complementary, with each playing a crucial role in the SubjectDrive framework. Third, the final model includes ATA, complementing both SVA and SPA, and yields further improvements of 0.7 in AMOTA and 1.0 in RECALL. These results underscore the efficacy of ATA in enhancing alignment accuracy and temporal consistency across tasks.

\begin{table}[!h]
    \centering
    \small
     \renewcommand{\arraystretch}{1.0}
         \setlength{\tabcolsep}{0.1pt}
        \begin{tabular}{cccccc}
            \toprule
            SVA &SPA   &ATA  &NDS$\uparrow$  & AMOTA$\uparrow$ &RECALL $\uparrow$   \\
            \hline
             $-$  &  $-$  &  $-$ 	 & 32.1 & 11.4  & 21.3 \\
            $\checkmark$  &  $-$  &  $-$  	  & 34.3  & 15.6 & 24.5 \\
             $\checkmark$  &$\checkmark$ &  $-$   & 36.1  & 17.4  & 27.1 \\
            $\checkmark$  &$\checkmark$ & $\checkmark$  & 36.3 (+\textbf{4.2\%}) & 18.1 (+\textbf{6.7\%}) & 28.1 (+\textbf{6.8\%})  \\
            \hline
              & Real Validation   & & 46.9 & 30.1  & 41.8 \\
            \bottomrule
        \end{tabular}
     \caption{
    Ablation studies of different modules in SubjectDrive, with the last row showing the alignment performance on the real validation data.
     }
        \label{tab:ablation}
\end{table}

\subsection{Visualisation results}

\begin{figure}[!h]
  \centering
   \includegraphics[width=0.815\linewidth]{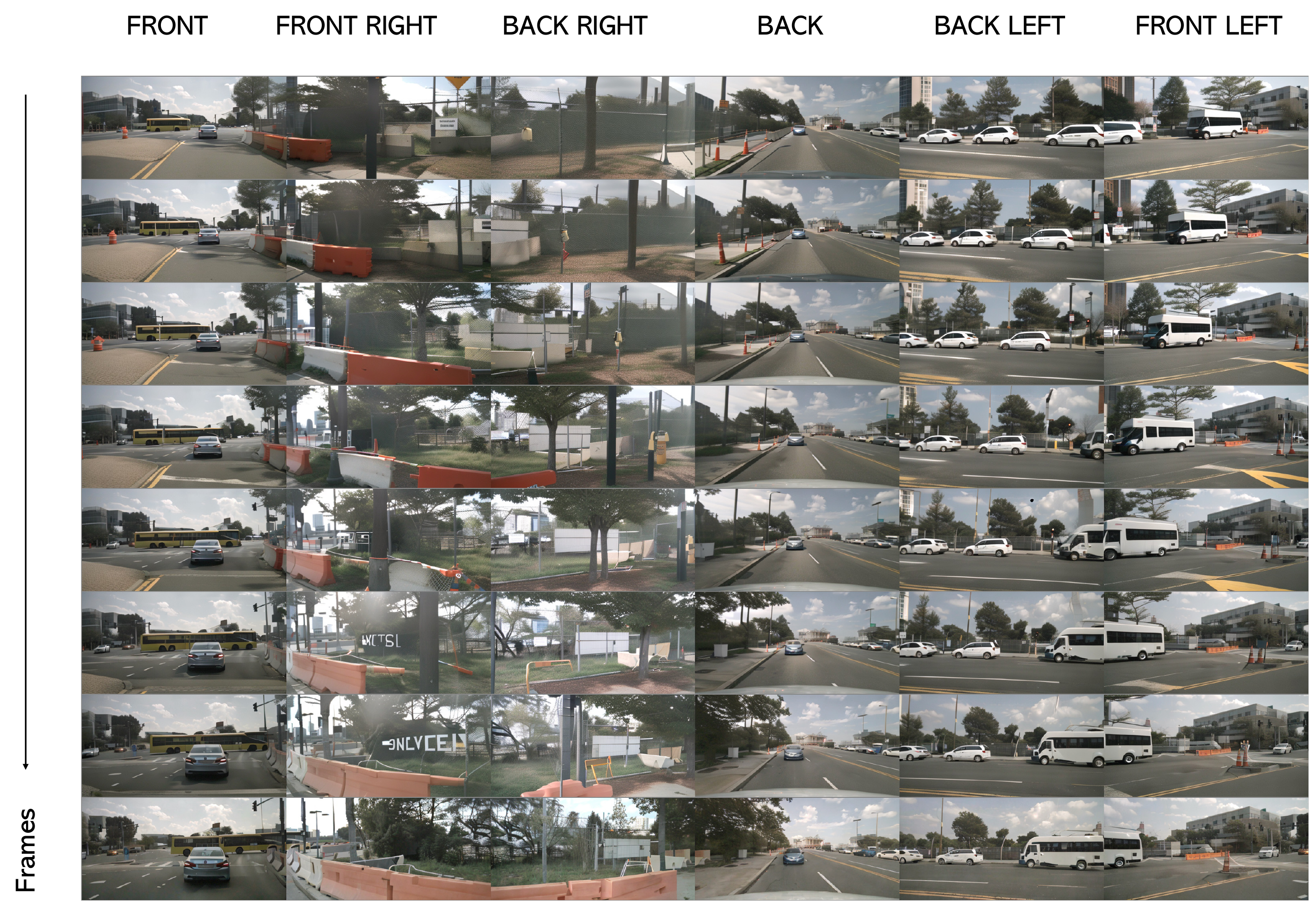}
   \hfill
   \caption{Multi-view videos generated by Ours.}
   \label{fig:tconsis}
\end{figure}

\noindent\textbf{Consistent Multi-View Video Generation.}
As illustrated in Fig. \ref{fig:tconsis}, for the six-view, eight-frame generated video, SubjectDrive produces temporally and view-consistent videos on the nuScenes validation set.

\noindent\textbf{Subject-controlled Video Generation.} Fig. \ref{fig:subject} shows the visualization of subject-controlled driving video generation. Given the image of a reference subject, SubjectDrive can generate layout-aligned driving videos featuring the desired subject. By using reference subjects as control signals, SubjectDrive offers a mechanism for incorporating external diversity into the generated data.
\begin{figure}[!t]
  \centering
   \includegraphics[width=0.83\linewidth]{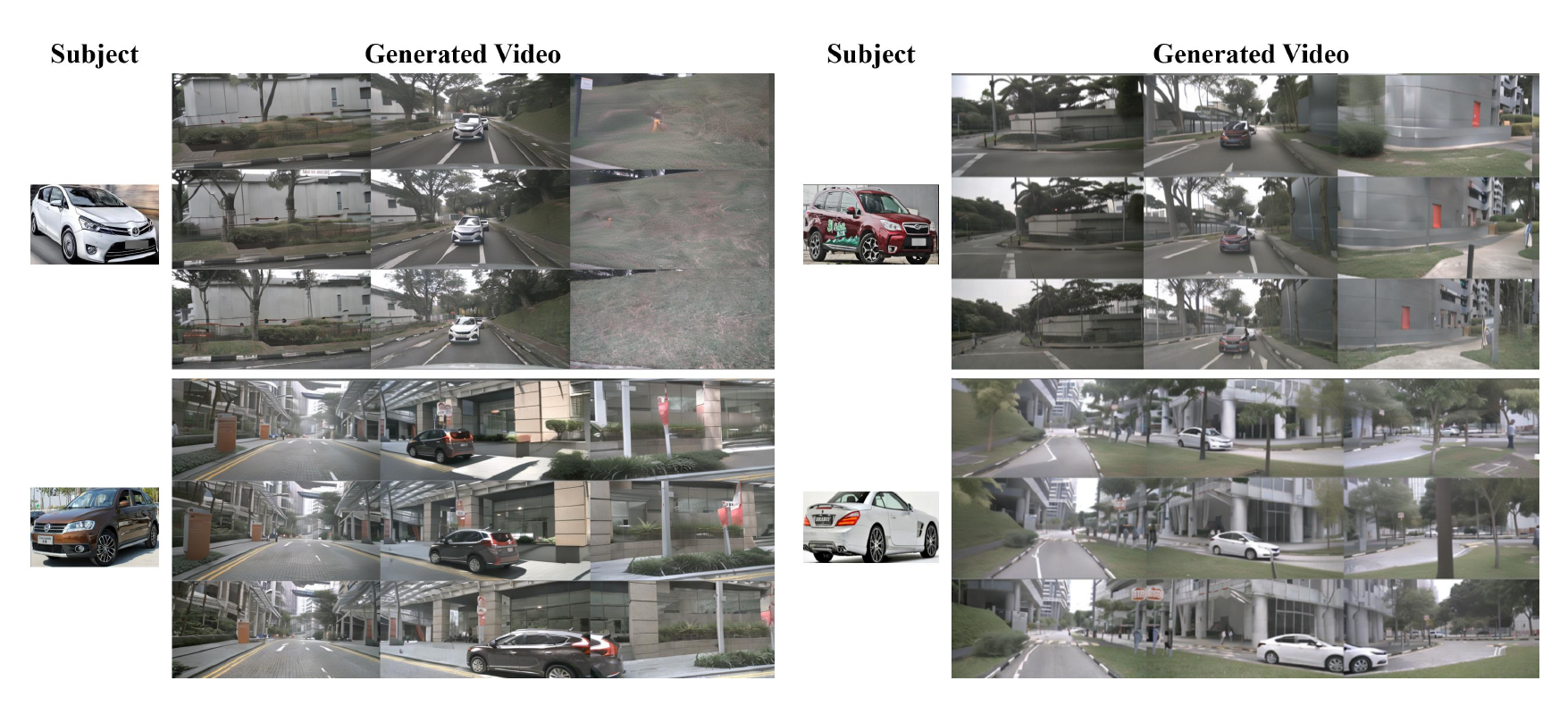}
   \hfill
   \caption{Subject-controlled videos generated by Ours.}
   \label{fig:subject}
\end{figure}

\begin{figure}[!h]
  \centering
   \includegraphics[width=0.835\linewidth]{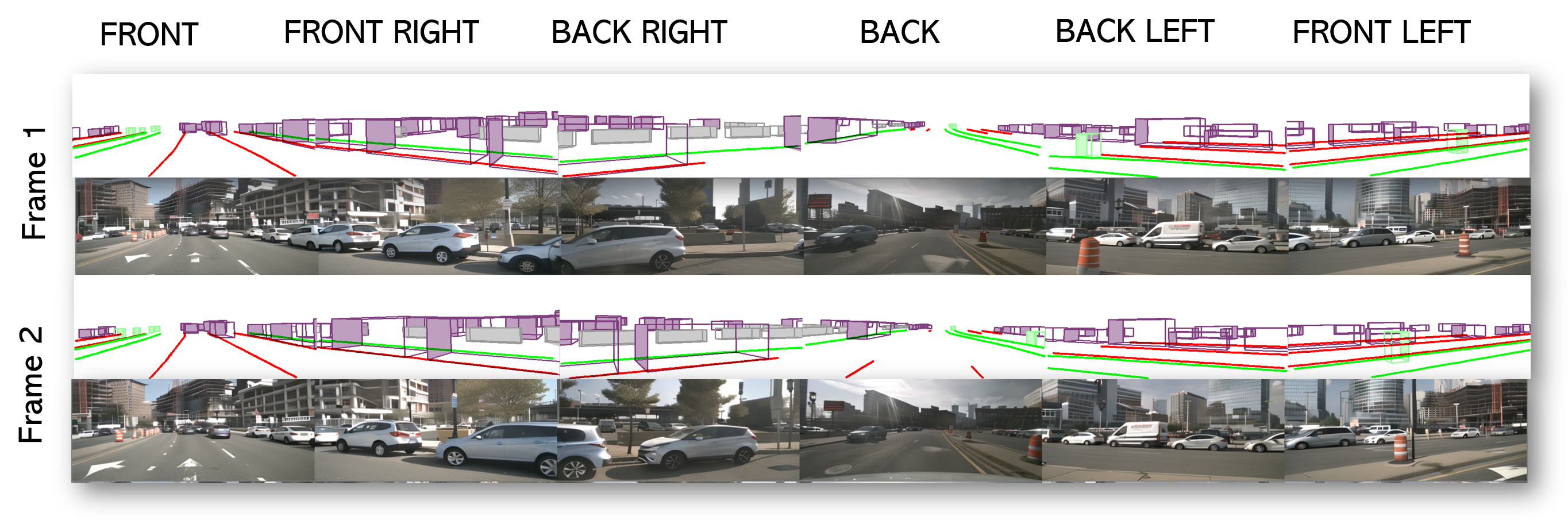}
   \hfill
   \caption{Controllable multi-view videos generated by Ours.}
   \label{fig:alignbev}
\end{figure}

\noindent\textbf{Controllable Video Generation.}
Fig. \ref{fig:alignbev} illustrates the driving scene video generated by our method, which adheres to the BEV layout conditions. From this visualization, it is evident that our generated synthetic data closely aligns with the specified BEV conditions, showcasing superior layout control and alignment capabilities.

\section{Conclusion}

In this work, we present SubjectDrive, a novel video generation framework that enhances the scalability and sampling diversity of generative models. The architecture incorporates three key innovations: a subject prompt adapter, a subject visual adapter, and augmented temporal attention, collectively enabling robust subject control. This feature significantly improves the model's ability to generate diverse samples. Extensive experiments demonstrate that SubjectDrive not only outperforms existing methods but also scales effectively. Notably, it is the first generative model to enhance perception performance beyond pre-trained capabilities on the nuImages dataset, showcasing the transformative potential of generative data in advancing autonomous driving technologies and pointing to promising future directions in the field.

\section{Acknowledgments}
The work was supported by National Science and Technology Major Project of China (2023ZD0121300). 

\bibliography{aaai25}

\end{document}